\documentclass[runningheads]{llncs}
\usepackage{graphicx}
\usepackage{subfig}
\usepackage{algorithm}
\usepackage{algorithmic}
\usepackage{amsmath}
\usepackage{amssymb}
\usepackage{booktabs}
\usepackage{wrapfig}
\begin{document}
\title{Robustness Threats of Differential Privacy}
%
%

\author{Nurislam Tursynbek\textsuperscript{\rm 1}, Aleksandr Petiushko\textsuperscript{\rm 2,3}, Ivan Oseledets\textsuperscript{\rm 1}
}
\institute{\textsuperscript{\rm 1}Skolkovo Institute of Science and Technology, \textsuperscript{\rm 2}Huawei MRC, \textsuperscript{\rm 3}Lomonosov MSU\\
{\tt\small nurislam.tursynbek@gmail.com, petyushko.alexander1@huawei.com, i.oseledets@skoltech.ru}
}

\authorrunning{Tursynbek et al.}
%
%
\maketitle              
\begin{abstract}
Differential privacy (DP) is a gold-standard concept of measuring and guaranteeing privacy in data analysis. It is well-known that the cost of adding DP to deep learning model is its accuracy. However, it remains unclear how it affects robustness of the model. Standard neural networks are not robust to different input perturbations: either adversarial attacks or common corruptions. In this paper, we empirically observe an interesting trade-off between privacy and robustness of neural networks. We experimentally demonstrate that networks, trained with DP, in some settings might be even more vulnerable in comparison to non-private versions. To explore this, we extensively study different robustness measurements, including FGSM and PGD adversaries, distance to linear decision boundaries, curvature profile, and performance on a corrupted dataset. Finally, we study how the main ingredients of differentially private neural networks training, such as gradient clipping and noise addition, affect (decrease and increase) the robustness of the model.

\end{abstract}

\section{Introduction}
Although neural networks achieved state-of-the-art performance in many problems, they are vulnerable to different malicious attacks. Adversaries might harm the systems by breaking the robustness of the neural network (security risks) or by inferring the secret information (privacy risks). These risks endanger their deployment in some specific environments, such as medical image analysis \cite{gomathisankaran2013ensure} and video surveillance \cite{rajpoot2014security}, where models have to be both secure and private. 

Differential privacy (DP) \cite{dwork2008differential} is a powerful concept of guaranteeing privacy of data. 
It provides a promise claiming that training data will not be affected, adversely or otherwise, no matter what other information is available elsewhere \cite{dwork2008differential}. This definition is not too strict, as surprisingly anonymous data can be robustly de-anonimized using other publicly available information, even for large sparse datasets \cite{narayanan2008robust,aggarwal2005k}. Moreover, to preserve privacy of one training sample, we should assume that the adversary has access to some or even all other training samples. This scenario is not unrealistic, and can be observed when the data is crowdsourced. To mitigate this, Abadi et al. \cite{abadi2016deep} proposed DP-SGD to apply DP to the learning procedure. This method protects individual privacy from strong adversaries with the full knowledge of the training procedure and model's hyperparmeters, and learns useful information about the dataset as a whole.

\begin{figure}[t]
  \centering
  \hspace{0.3cm}
  \subfloat[Trained with SGD]{\includegraphics[width=0.3\linewidth]{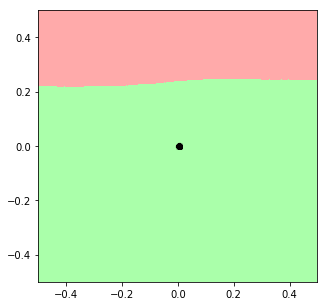}\label{fig:f1}}
  \hspace{0.3cm}
  \subfloat[Trained with DP-SGD]{\includegraphics[width=0.3\linewidth]{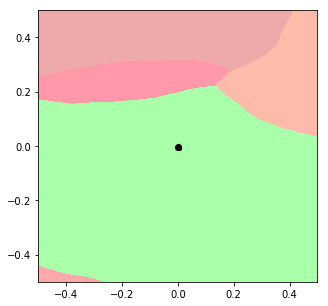}\label{fig:f2}}
  \hspace{0.3cm}
  \caption{Random normal cross-sections of the decision boundary for CNN classifiers trained with SGD and DP-SGD on MNIST. The green region is correct class red regions are incorrect classes. The point at the center shows the sample. The classification regions are shown in the plane spanned by normal to the decision boundary (y-axis) and a random orthogonal direction (x-axis). Differential Privacy (DP-SGD) affects the underlying geometry of the decision boundaries.
}
\label{fig1}
\end{figure}

On security side, researchers discovered intriguing properties of neural networks \cite{szegedy2013intriguing,biggio2013evasion} of being brittle to imperceptibly perturbed inputs that fool neural networks, known as adversarial examples \cite{goodfellow2014explaining}. FGSM is a fast adversarial attack that maximally increases loss function in a single step \cite{goodfellow2014explaining}. Several other popular approaches to find adversarial examples are described in Section \ref{measurements}. Furthermore, adversarial examples are also found in physical world \cite{kurakin2016adversarial} in safety-critical areas, such as self-driving cars \cite{eykholt2018robust}, face identification systems \cite{sharif2016accessorize}, speech recognition technologies \cite{carlini2018audio}. These issues brought serious security risks of using deep learning in the real world, compromising people's confidence in neural networks.

Several defenses that make neural networks robust or privacy-preserving were proposed recently. However, a typical defense concentrates in one domain: either privacy-preservance or robustness. This still makes it unshielded and obscure in another. For instance, it is unclear how training procedure that guarantees privacy (DP-SGD \cite{abadi2016deep}) might affect the robustness. As an example how DP impacts a neural network, the comparative illustration of normal cross-sections of the decision boundaries for differentially private and naturally trained classifiers is shown in Fig.\ref{fig1}. The notable difference suggests that defending from privacy attacks also affect the underlying decision boundaries and robustness of model.

DP-SGD enhances privacy by clipping per-example gradients and adding Gaussian noise to each update steps. The objective is to make the model invariant to the presence or absence of each training sample, making their impact less observable. Thus, intuitively, DP have the potential to equalize (decrease or increase) the "conspicuousness" or surrounding area (such as $\ell_p$ ball) of samples for which model prediction is unchanged, making the model less or more robust.   

In this work, we experimentally reveal the limitation of focusing solely on defending from privacy attacks with DP, and its impact to the robustness. 

\section{Background}


\subsection{Differential Privacy}
Differential privacy (DP) \cite{dwork2008differential} quantifies how much information is leaked during a particular mechanism and restricts the impact of each sample, by introducing the degree of randomness to the algorithm as a part of its logic. 
\newtheorem{dfn}{Definition}

\begin{dfn}[Differential Privacy]
A randomized algorithm (or mechanism) $\mathcal{A}: \mathcal{D}\rightarrow\mathcal{R}$  with domain $\mathcal{D}$ and range $\mathcal{R}$ gives $(\mathcal{E}, \delta)-$differential privacy ($\mathcal{E},\delta>0$), if for all datasets $D\in\mathcal{D}$ and $D'\in\mathcal{D}$, differing in at most one element, and for any subset of possible outputs $\mathcal{S}\subseteq\mathcal{R}$
, it is true that:
$$Pr[\mathcal{A}(D)\in\mathcal{S}]\leq e^{\mathcal{E}} Pr[\mathcal{A}(D')\in\mathcal{S}]+\delta$$
\end{dfn}

Here, $\mathcal{E}$ is a privacy budget and $\delta$ (generally, $\delta<1/|D|$) is the probability of possible violation of $\mathcal{E}$-DP. The smaller the $\mathcal{E}$, the more private the model.

For neural networks, the algorithm $\mathcal{A}$ is the training procedure, which takes dataset $D$ as input, learns from it with Stochastic Gradient Descent (SGD), and returns a model as output. To make this mechanism private, Abadi et al. \cite{abadi2016deep} introduced an elegant approach called differentially private SGD (DP-SGD).

\subsection{Differentially Private SGD \cite{abadi2016deep}}
\begin{wrapfigure}{r}{0.54\textwidth}
\vspace{-1.5cm}
\begin{minipage}{0.54\textwidth}
\begin{algorithm} [H]
        \caption{DP-SGD \cite{abadi2016deep}}
        \label{alg:ALG1}
        \textbf{INPUT:} Dataset $\{(x_i, y_i)\}$ batch size $b$, learning rate $\eta$, loss $\mathcal{L}(\theta(x), y)$, $T$ iterations, noise $\sigma$, clipping bound $C$
        \begin{algorithmic}
        \STATE  \textbf{Initialize} $\theta_0$ randomly
        \FOR  {$t=1$ to $T$}
        \STATE Sample random batch $B_t$ of size $b$ 
        \FOR {$i=1$ to $b$}
        \STATE $g_t(x_i)=\nabla\mathcal{L}_{\theta}(\theta, x_i)$
        \STATE $\overline{g}_t(x_i)=\frac{g_t(x_i)}{\max\left(1,\frac{\|g_t(x_i)\|_2}{C}\right)}$
        \ENDFOR
        \STATE $\Tilde{g}_t = \frac1b\left(\sum\limits_{x_i\in B_t}\overline{g}_t(x_i)+\mathcal{N}(0, \sigma^2C^2\textbf{I})\right)$ 
        \STATE $\theta_{t+1}=\theta_t - \eta\Tilde{g}_t$
        \ENDFOR
        \end{algorithmic}
        \textbf{OUTPUT:} Model $\theta_T$
\label{alg1}
\end{algorithm}
\end{minipage}
\vspace{-0.8cm}
\end{wrapfigure}
To update weight $\theta$ of a neural network by minimizing the loss function $\mathcal{L}$, standarad SGD approximates the gradient direction $-\nabla_{\theta}\mathcal{L}_{\theta}$ using a small, randomly chosen portion of it called batch. Usually, to make an algorithm random, the noise mechanism is added to the input, output, or the intermediate parts of the algorithm. In the case of SGD, adding noise to the input $X$ or the model's parameters $\theta$ (output of the training procedure) would destroy the utility of learned model. To make the learning algorithm SGD differentially private, gradient noise was proposed \cite{abadi2016deep}. Before noise addition, each individual component of the gradient is norm-clipped to bound the influence of each data point. Due to the lack of prior information about the norm of the gradients, each gradient $\nabla\mathcal{L}_{\theta}(\theta, x_i)$ is clipped by $\ell_2$-norm, upper-bounded by threshold $C$. The full batch gradient is computed as a Gaussian noise with zero mean vector and covariance matrix $\sigma^2C^2\mathbf{I}$  added to the sum of each individual gradients, divided by the size of a batch. Procedure is summarized in Algorithm \ref{alg1}.

    
\subsection{Robustness Measurements to Adversarial Examples and Common Corruptions}
\label{measurements}
Adversarial examples are imperceptibly perturbed inputs, that completely fool neural networks. These intriguing properties of neural networks were discovered in \cite{szegedy2013intriguing}, by solving an optimization problem for a given input sample $x$:
\begin{equation}
\min_{\Delta}\|\Delta\|_p\quad s.t.\;\; \arg\max_{i}\mathcal{F}_{\theta}(x+\Delta)_{i}\neq\arg\max_{i}\mathcal{F}_{\theta}(x)_{i},
\end{equation}
where $\mathcal{F}_{\theta}(\cdot)_{i}$ is the $i$-th output class of a classifier with parameters $\theta$. 

\subsubsection{FGSM}
Goodfellow et al. \cite{goodfellow2014explaining} proposed an efficient single-step way to compute an $\ell_{\infty}-$bounded perturbation called Fast Gradient Sign Method (FGSM):
\begin{equation}
x_{adv} = x + \varepsilon\; sign(\nabla_x \mathcal{L}(\theta,x,y)),
\end{equation}
where $y$ is the true class, $\nabla_x\mathcal{L}(\theta,x,y)$ computes the gradient of a cost function with respect to $x$. FGSM is a simple and effective way of testing robustness.

\subsubsection{PGD}
A straightforward iterative extension of FGSM was found to be successful in \cite{kurakin2016adversarial,madry2017towards}. Madry et al. \cite{madry2017towards} indicated its equivalence to $\ell_{\infty}$ version  of constrained optimization method, called Projected Gradient Descent (PGD). PGD begins with $x_{adv}^0 = x$ and then iteratively changes with a step-size $\alpha$:
\begin{equation}
x_{adv}^{t+1} = Clip\left\{x_{adv}^t + \alpha sign(\nabla_x \mathcal{L}(\theta, x_{adv}^t,y)), \; x-\varepsilon, x+\varepsilon \right\}
\end{equation}
After each update, the input is also clipped to pixel constraints. PGD adversarial samples are the main tool to construct adversarially trained models \cite{madry2017towards}.

\subsubsection{DeepFool}
DeepFool \cite{moosavi2016deepfool} is a simple and accurate few-step method to compute a smaller adversarial perturbation to the input. It finds a distance to the closest locally linearized decision boundary, which makes input to become an adversarial example. The closest $\ell_2$ and $\ell_\infty$ distances from $x$ with a $true$ class to a decision boundary of any class $k$ are approximated by:
\begin{equation}
\|l_k(x)\|_2=\frac{|f_{true}(x)-f_k(x)|}{\|\nabla_x f_{true}(x)-\nabla_x f_k(x)\|_2}; \quad \|l_k(x)\|_\infty=\frac{|f_{true}(x)-f_k(x)|}{\|\nabla_x f_{true}(x)-\nabla_x f_k(x)\|_1}
\label{eq5}
\end{equation}
Here, $f_k(x)$ is the output of predicted logits, corresponding to the $k-$th class and $\nabla_x f_k(x)$ is its gradient with respect to input $x$. The closest distance to wrong (adversarial) class is computed as $\min\limits_{k\neq true} l_k(x)$. 
\subsubsection{CURE}
Moosavi et al. \cite{moosavi2019robustness} discovere that curvature of the boundaries of a classifier is inversely proportional to its robustness. They found that decision boundaries of robust models are less curved and proposed to CUrvature REgularization (CURE) as an alternative to the adversarial training. The curvature is suggested to compute as eigenvalues of a Hessian matrix:
\begin{equation}
H = \left(\frac{\partial^2\mathcal{L}}{\partial x_i\partial x_j}\right)\in \mathbb{R}^{d\times d}
\label{cureeq}
\end{equation}

\subsubsection{MNIST-C} A corrupted version of MNIST, known as MNIST-C \cite{mu2019mnist}, serves as an out-of-distribution robustness benchmark for computer vision models. Dataset contains $31$ different common corruptions applied to the standard MNIST, while preserving semantic content of images, however, significantly degrades the accuracy of a model, even adversarially robust models, showing that adversarial robustness is not transferable to robustness against common corruptions.

\section{Experimental study}
\label{Experiments}
Here, we describe the experimental evaluations of robustness for models trained with SGD and DP-SGD. Following \cite{abadi2016deep}, we perform experiments on MNIST and CIFAR-10 dataset, since its easier and fast to train models on it. This datasets are public and serve as a benchmark for different machine learning tasks. In all experiments we use Pytorch framework and NVidia GeForce GTX GPUs.

For MNIST, which contains $60000$ training and $10000$ testing gray-scale images of size $28\times28$, we used a LeNet architecture which consists of two convolutional layers with kernel size $5\times5$, stride $1$, and number of channels $20$ and $50$, with $2\times2$ max-pooling, following by two fully-connected layers with $500$ hidden units each. For all layers, ReLU activation is used. Learning rate is $0.05$, number of epochs is $50$ epochs, batch size is $256$. In subsections \ref{RMPGD}, \ref{RMDCDB}, \ref{curesection} and \ref{RMCC}, the noise $\sigma=1.3$ and the clipping bound $C=1.0$ are used. 

\subsection{Robustness Measurement with FGSM}
\begin{wrapfigure}{r}{0.5\textwidth}
\vspace{-0.7cm}
\includegraphics[width=0.5\textwidth]{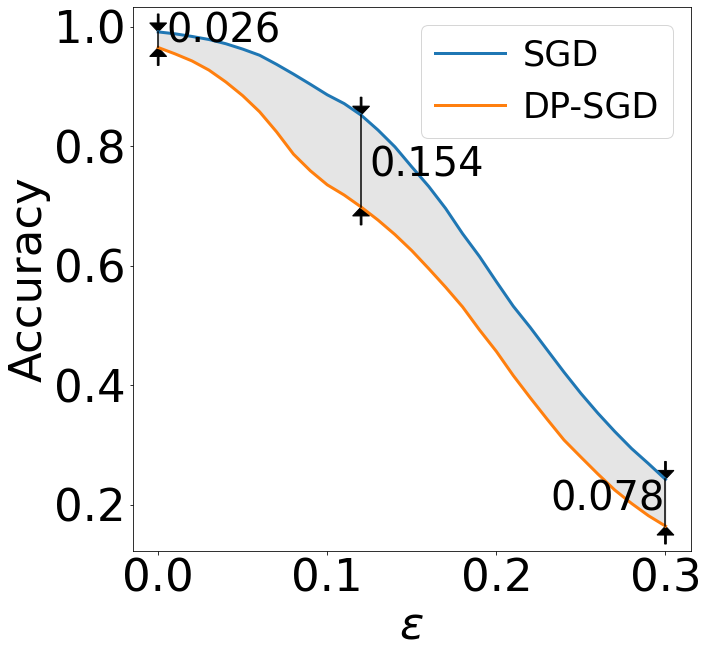}
\caption{Decrease of accuracy with increasing strength of FGSM adversary for neural networks trained with SGD and DP-SGD. Here $\varepsilon$ is adversary's strength ($\ell_{\infty}$ of perturbation) for MNIST}
\label{fig2}
\vspace{-0.7cm}
\end{wrapfigure}
FGSM significantly degrades the accuracy of models depending on the $\ell_{\infty}$-bound of the perturbation $\varepsilon$, showing neural networks trained with standard SGD are not robust to simple adversarial example. Surprisingly, the models trained with DP-SGD can be even more non-robust, i.e. the accuracy for these models decrease quicker. Figure \ref{fig2} demonstrates the plots of accuracies depending on perturbation $\ell_{\infty}$-norm $\varepsilon$. One might expect the accuracy drop would be similar for different training procedures, but DP-SGD decreases faster, increasing initial gap $2.6\%$ between accuracies, sometimes, reaching $15.4\%$ for $\varepsilon=0.12$, and always more than initial value. More swift accuracy drop indicates that DP-SGD reduces robustness to FGSM adversarial examples. 

\begin{figure}[H]
  \centering
  \subfloat[Dependence on $\sigma$]{\includegraphics[width=0.33\textwidth]{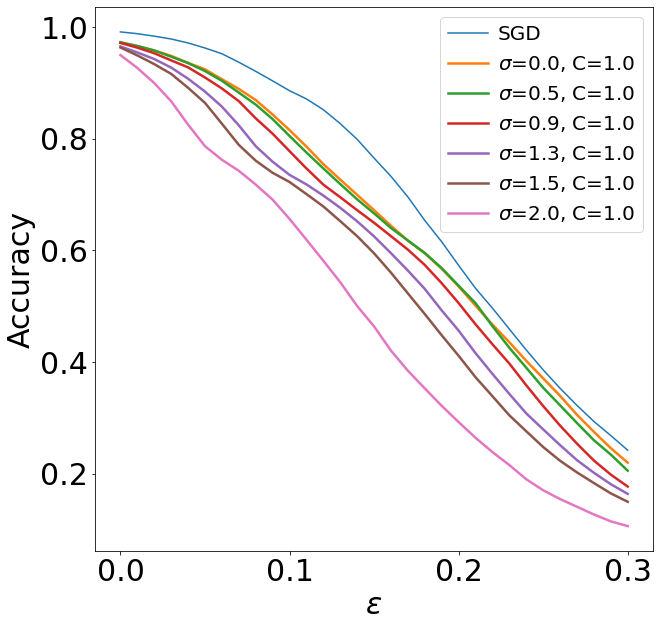}\label{noisa}}
  \hfill
  \subfloat[Dependence on $C$]{\includegraphics[width=0.33\textwidth]{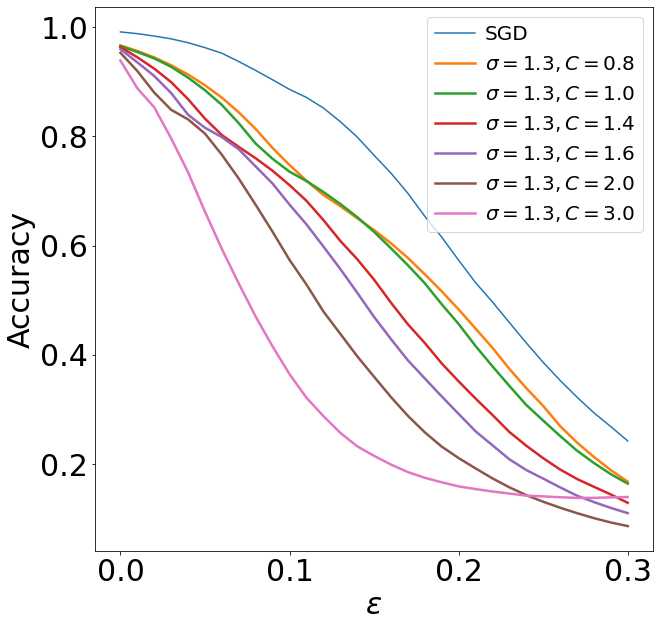}\label{noisb}}
  \hfill
  \subfloat[Dependence on $C$, when noise is $\mathcal{N}(0, \sigma^2\textbf{I})$ ]{\includegraphics[width=0.33\textwidth]{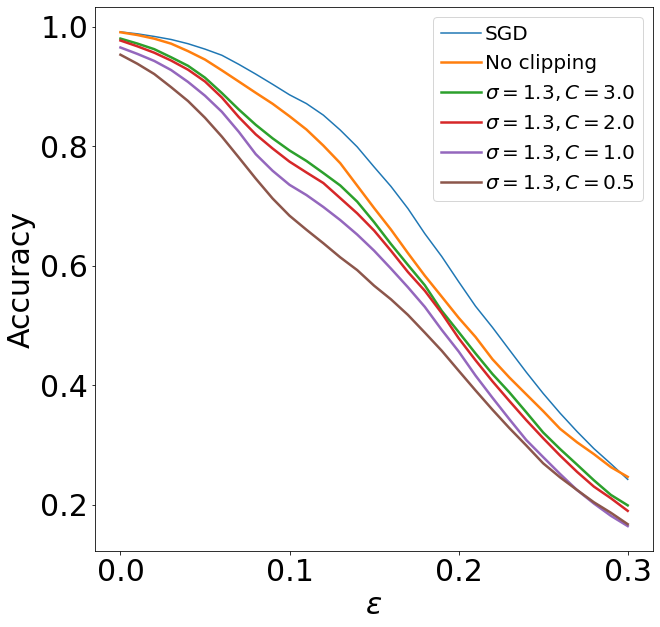}\label{noisc}}
  \caption{FGSM accuracies for different  noises and clipping bounds for MNIST.
}
\label{noiseclip}
\end{figure}

For DP-SGD, we used noise $\sigma=1.3$ and the clipping bound $C=1.0$. Based on \cite{abadi2016deep}, these values give privacy budget $\mathcal{E}<5$ for $\delta=10^{-6}$. Next, we explore the dependence of two main ingredients of DP-SGD: gradient clipping and noise addition. In Fig.\ref{noiseclip} we show the dependence of accuracy curves for different $\ell_{\infty}$ of perturbation $\varepsilon$ for models with varying noise $\sigma$ and clipping bound $C$.

In Figure \ref{noisa}, the gap is increasing faster with the increase of the noise $\sigma$, meaning more private (noisy) algorithms are less robust. Figure \ref{noisb} shows the dependence of FGSM curves on the clipping bound $C$. Since $C$ also appears in the noise $\mathcal{N}(0, \sigma^2C^2\textbf{I})$, it also decreases robustness by increasing the accuracy gap even faster. Therefore, we explore how pure clipping affects robustness. For this, we vary $C$ only for clipping the norm of gradients, while keeping the noise as $\mathcal{N}(0, \sigma^2\textbf{I})$ , not $\mathcal{N}(0, \sigma^2C^2\textbf{I})$. The result in Figure \ref{noisc} indicate that private models that bounds individual gradients to smaller values are less robust.

\subsection{Robustness Measurement with PGD}
\label{RMPGD}

\begin{wrapfigure}{r}{0.6\textwidth}
\vspace{-1.9cm}
\begin{minipage}{0.6\textwidth}
\begin{table}[H]
\caption{PGD Results for MNIST.}
  \begin{tabular}{cccc}
    \toprule
    PGD settings & SGD Acc.& DP-SGD Acc. &Difference\\
    \midrule
    $\epsilon=0.3, steps=40$ & 1.00\% & 0.23\% & 0.77\% \\
    $\epsilon=0.2, steps=40$ & 10.15\% & 4.36\% & 5.79\%\\
    $\epsilon=0.1, steps=40$ & 80.44\% & 60.89\% & 19.55\%\\
    $\epsilon=0.3, steps=30$ & 3.06\% & 1.38\% & 1.68\%\\
    $\epsilon=0.2, steps=30$ & 15.84\% & 7.88\% & 7.96\%\\
    $\epsilon=0.1, steps=30$ & 80.63\% & 61.10\% & 19.53\%\\
    \bottomrule
\end{tabular}
\label{tablepgd}
\end{table}
\end{minipage}
\vspace{-0.9cm}
\end{wrapfigure}

As mentioned in \cite{madry2017towards}, the PGD finds local maxima of the loss function, based on the constrains the adversary has, and serves as “universal” adversary among first-order approaches. We can use this information to compare robustness of a neural network.

First, we compare accuracy of the network when PGD attack is applied for a fixed number of steps and adversary's strength $\varepsilon$. The results are shown in Table \ref{tablepgd} and indicate that for $\varepsilon=0.1;0.2$, the gap between SGD and DP-SGD increases compared to initial gap $2.6\%$, which indicates that DP-SGD trained network can be less robust to adversaries with small strength.

\begin{figure*}[t]
  \centering
  \subfloat[$\alpha=0.005$]{\includegraphics[width=0.25\textwidth]{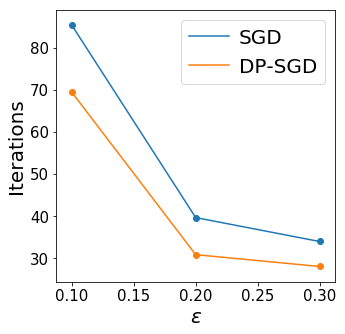}\label{fig:f1}}
  \hfill
  \subfloat[$\alpha=0.01$]{\includegraphics[width=0.25\textwidth]{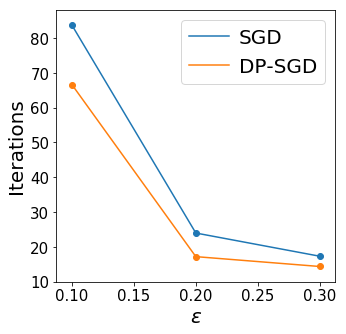}\label{fig:f1}}
  \hfill
  \subfloat[$\alpha=0.02$]{\includegraphics[width=0.25\textwidth]{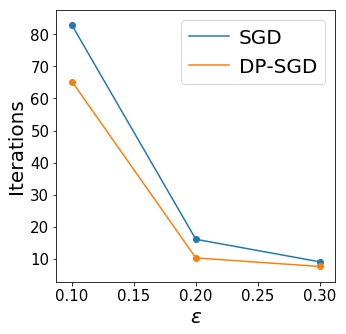}\label{fig:f1}}
  \hfill
  \subfloat[$\alpha=0.05$]{\includegraphics[width=0.25\textwidth]{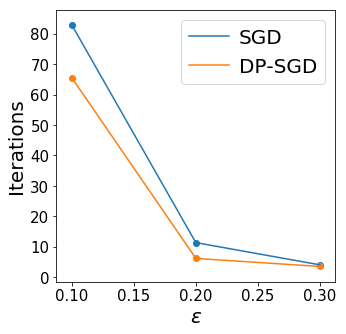}\label{fig:f1}}
  \hfill
  \caption{Number of PGD iterations, needed to achieve adversarial class for MNIST.
}
\label{pgditer}
\end{figure*}

Next, we compare a number of iterations to reach an adversarial class for a particular adversarial strength $\varepsilon$. The comparison is shown in Figure \ref{pgditer} for different step-sizes $\alpha$, in case when adversarial strength $\varepsilon=0.1;0.2;0.3$. For the small $\varepsilon$, it might be the case that adversary does not find adversarial class. Less number of iterations needed for PGD to fool the network with DP-SGD training indicates that DP-SGD might reduce robustness to PGD adversaries.

\subsection{Robustness as Distance to Closest Decision Boundary}
\label{RMDCDB}

In \cite{moosavi2016deepfool} authors find a perturbation as a vector, directed to the closest linearly approximated decision boundary. To calculate $\ell_2$ and $\ell_{\infty}$ distances, we use \eqref{eq5} for all test samples. The histograms of distances are in Figure \ref{deepfool1} and \ref{deepfool2}. 

In Figures \ref{deepfool1} and \ref{deepfool2}, the decision boundaries of the model trained with DP-SGD are closer to data samples. For DP-SGD, the average value for $\ell_{\infty}-$bounded closest distance is reduced from $0.154$ to $0.131$ ($\sim 15\%$ drop) and $\ell_{2}-$bounded closest distance is reduced from $1.965$ to $1.654$ ($\sim16\%$ drop). Distance reduction indicates that DP-SGD builds a boundary that is close to data samples, which makes the model more susceptible to DeepFool adversaries, or less robust.
\label{deepfoolsection}

\subsection{Robustness Measurement as Curvature Profile}
\label{curesection}

The random normal cross-section of the decision boundaries for a CNN classifier in Figure \ref{fig1} shows that the decision boundaries of the classifier trained with DP-SGD "look more curved" compared to the the boundaries of the classifier trained with SGD. According to the recent finding in \cite{moosavi2019robustness}, more robust models have less curved (more 'linear') decision boundaries, and vice versa. 

\begin{figure*}[t]
  \centering
  \subfloat[Histogram of distances to the closest decision boundary for MNIST in $\ell_\infty$]{\includegraphics[width=0.33\textwidth]{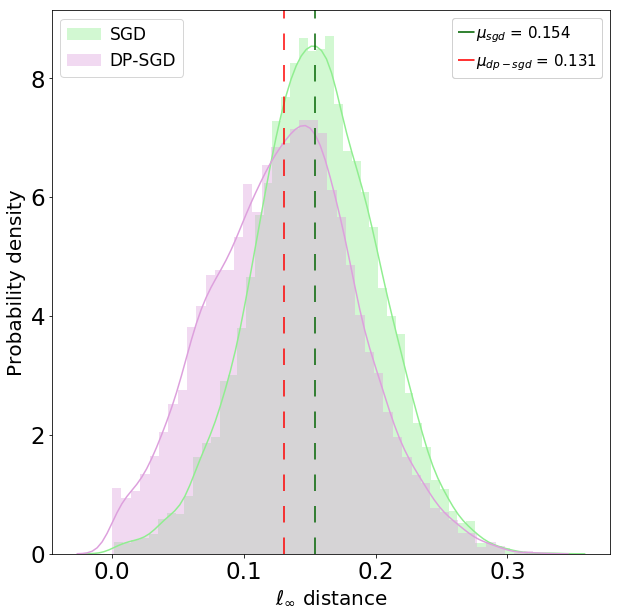}\label{deepfool1}}
  \hfill
  \subfloat[Histogram of distances to the closest decision boundary for MNIST in $\ell_2$]{\includegraphics[width=0.33\textwidth]{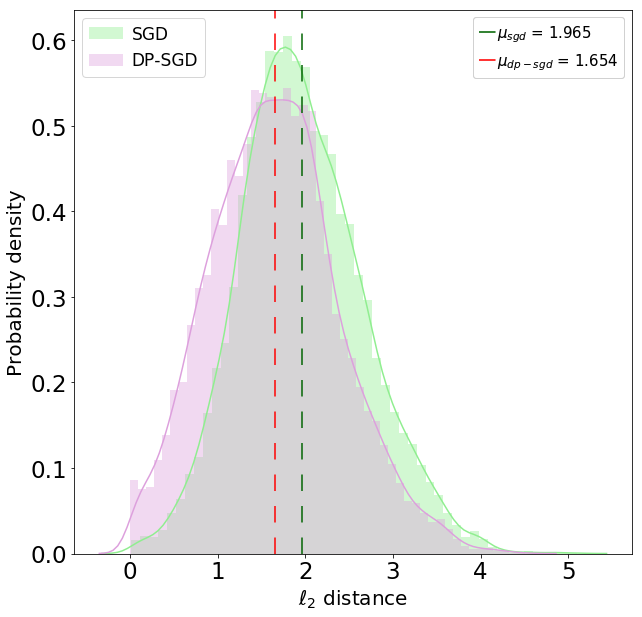}\label{deepfool2}}
  \hfill
  \subfloat[Curvature profiles as largest 15 eigenvalues of Hessian matrix for MNIST. ]{\includegraphics[width=0.33\textwidth]{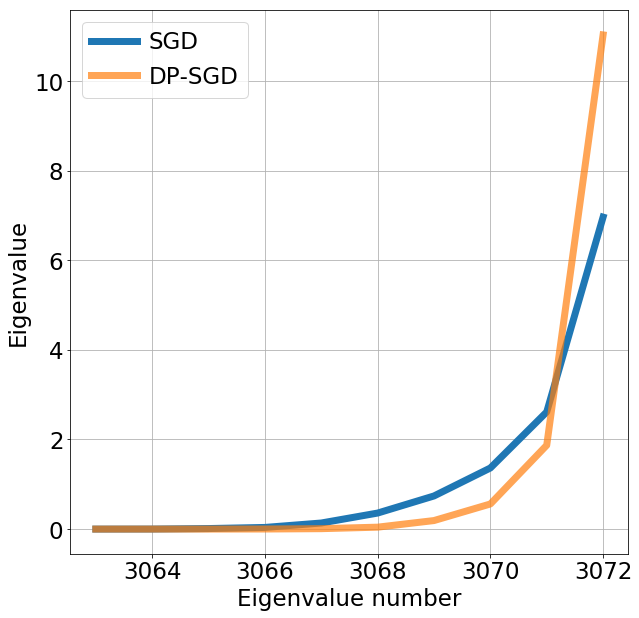}\label{cure2}}
  \caption{Distances and curvature results of MNIST experiments.}

\end{figure*}

To quantitatively measure the curvature of the network, following \cite{moosavi2019robustness}, we compute the curvature profile as sorted eigenvalues of Hessian matrix \eqref{cureeq} at $100$ random test samples. Since most of the eigenvalues are close to $0$, to better understand curvature profiles, we plot a logarithmic-scaled largest $15$ eigenvalues in Figure \ref{cure2}. Curvature profile of the model trained with DP-SGD has higher eigenvalue magnitudes (the largest eigenvalue difference is $100$ times) indicating that the model trained with DP-SGD produces more curved decision boundaries, and thus makes it less robust to adversarial examples \cite{moosavi2019robustness}.

\subsection{Robustness Measurement to Common Corruptions}
\label{RMCC}

Common corruptions are also considered as a robustness benchmark. These corruptions were initially proposed in \cite{hendrycks2019benchmarking}, and the extension to MNIST was recently introduced in \cite{mu2019mnist}. Here, we describe the results of evaluation of neural networks trained with SGD and DP-SGD for the robustness benchmark MNIST-C.

The results are shown in Table \ref{tablemnist}. They clearly indicate that differential privacy has significant impact on model's robustness to common corruptions. The initial baseline difference gap between SGD and DP-SGD model of $2.61\%$ increases for almost all corruptions and in average it increases to $15.99\%$, sometimes reaching extremely large gaps of $69.09\%, 64.57\%, 63.47\%$. The largest gaps are produced for Brightness, Fog and Frost corruptions, which is interesting, because visually these three corruptions are also similar to each other. We also show evaluations on MNIST-C dataset for different values of DP-SGD parameters $\sigma$ and $C$. Results in Table \ref{tablemnist} indicate that increasing $C$ for the same $\sigma$ as well as increasing $\sigma$ for the same $C$ leads to increasing the accuracy drop between corrupted and not corrupted dataset.

\subsection{Experiments with CIFAR-10}

For CIFAR-10, which contains colour images of size $32\times32\times3$, similarly to \cite{abadi2016deep,koskela2018learning}, we used a neural network with two convolutional layers followed by three fully-connected layers. The convolutional layers use $3\times3$ convolutions with stride $1$, followed by ReLU and max-pooling of kernel size $3\times3$ with a stride $2$. The three fully-connected layers contain $500$ hidden units each, and ReLU activation. Since computing per-example gradients is computationally expensive, we pre-train convolutional layers on the CIFAR-100. Also, parameters of convolutional layers learned from one dataset often can be used on another without retraining on another \cite{jarrett2009best}. We train only the fully-connected layers for both SGD and DP-SGD. Learning rate is $0.01$, batch size is $200$ and number of epoch is $200$.

\begin{table}[H]
\caption{MNIST-C Results}
\hspace{-3cm}
  \begin{tabular}{cccc|||||cc|cc}
    \toprule
    Corruption & SGD & DP-SGD ($\sigma=1.3,C=1$) & Difference & $\sigma=1.3,C=0.8$ & $\sigma=1.3,C=1.2$ & $\sigma=0.7,C=1$ & $\sigma=1.5,C=1$ \\
    \midrule
    Baseline & 99.14 & 96.53 & 2.61 & 96.67 & 96.56 & 97.22 & 96.34\\
    \midrule
    Brightness & 96.71 & 27.62 & \textbf{69.09} &  32.22  &  22.54  &  36.43  &  23.5\\
    Fog & 91.59 & 27.02 & \textbf{64.57}&  29.68  &  24.8  &  33.12  &  24.95\\
    Impulse Noise & 95.34 & 89.43 & 5.91&  90.41  &  88.24  &  91.78  &  88.39\\
    Rotate & 92.77 & 84.34 & 8.43&  84.15  &  83.87  &  85.14  &  83.28\\
    Shot Noise & 98.06 & 95.09 & 2.97&  95.15  &  94.66  &  95.93  &  94.46\\
    Translate & 56.32 & 30.98 & 25.34&  30.74  &  30.32  &  33.32  &  29.25\\
    Canny Edges & 69.89 & 63.64 & 6.25&  65.89  &  62.04  &  67.62  &  62.6\\
    Glass Blur & 95.97 & 90.63 & 5.34&  90.82  &  90.32  &  91.91  &  90.12\\
    Scale & 94.98 & 71.72 & 23.26&  70.33  &  71.4  &  75.07  &  69.98\\
    Spatter & 97.12 & 94.13 & 2.99&  94.2  &  93.86  &  95.06  &  93.6\\
    Zigzag & 88.09 & 80.03 & 8.06&  80.37  &  79.46  &  82.33  &  79.08\\
    Dotted Line & 96.88 & 94.13 & 2.75&  94.07  &  93.51  &  94.97  &  93.42\\
    Motion Blur & 95.91 & 85.91 & 10.00&  84.7  &  86.18  &  87.29  &  85.4\\
    Shear & 97.84 & 91.16 & 6.68&  91.4  &  91.01  &  92.76  &  90.58\\
    Stripe & 88.05 & 62.39 & 25.66&  61.41  &  61.33  &  66.6  &  59.45\\
    Contrast & 95.57 & 71.66 & 23.91 &  77.5 & 65.79 & 84.34 & 66.61\\
    Defocus Blur & 95.66 & 88.96 & 6.70 & 88.94 & 88.97 & 90.05 & 88.77\\
    Frost & 94.05 & 30.58 & \textbf{63.47} & 33.98 & 27.84 & 38.66 & 28.11\\
    Gaussian Blur & 87.80 & 83.09 & 4.71 &82.81 & 83.09 & 82.12 & 83.39\\
    Gaussian Noise & 88.05 & 81.21 & 13.33& 82.69&79.94 & 85.07 & 80.16\\
    Pessimal Noise & 92.65 & 92.41 & 0.24 &93.18 &91.42 & 93.54 & 91.68\\
    Pixelate & 97.94 & 94.18 & 3.76 &94.19 & 93.93 & 95.42 & 93.62\\
    Saturate & 98.39 & 83.99 & 14.40 & 84.61 & 83.49 & 87.32 & 82.66\\
    Speckle Noise & 98.33 & 95.57 & 2.76 & 95.48 & 95.15 & 96.14 & 95.03\\
    Zoom Blur & 98.02 & 91.41 & 6.55 &91.6 & 91.25 & 93.12 & 90.94\\
    Jpeg Compression & 99.09 & 96.52 & 2.57 &96.29 & 96.32 & 97.05 & 96.1\\
    Elastic Transform & 86.73 & 72.0 & 14.73 & 71.31 & 71.65 & 74.0 & 70.78\\
    Quantize & 98.98 & 96.33 & 2.65 & 96.25 & 96.12 & 96.82 & 95.88\\
    Line & 85.56 & 82.53 & 3.03 & 82.6 & 81.71 & 84.16 & 81.45\\
    Inverse & 37.27 & 9.54 & 27.73 &9.20 &9.59 & 8.54 & 9.54\\
    Snow & 96.33 & 58.22 & 38.11 & 60.11 & 56.95 & 64.48 & 56.38\\
    \midrule
    Average & 90.72 & 74.73 & 15.99 & 75.36 & 73.77 & 77.42 & 73.52\\
  \bottomrule
\end{tabular}
  \label{tablemnist}
\end{table}

\begin{figure*}[t]
  \centering
  \subfloat[Histogram of distances to the closest decision boundary for CIFAR-10 in $\ell_\infty$]{\includegraphics[width=0.375\textwidth]{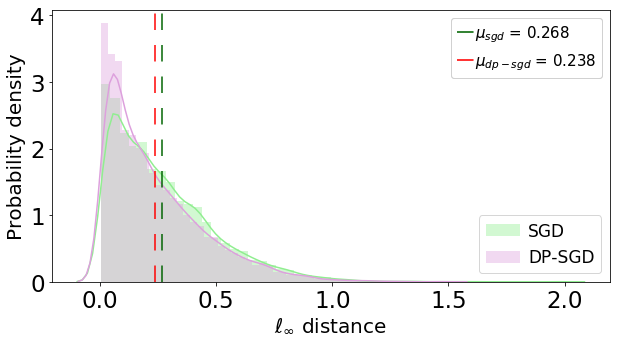}\label{deepcifarinf}}
  \hfill
  \subfloat[Histogram of distances to the closest decision boundary for CIFAR-10 in $\ell_2$]{\includegraphics[width=0.375\textwidth]{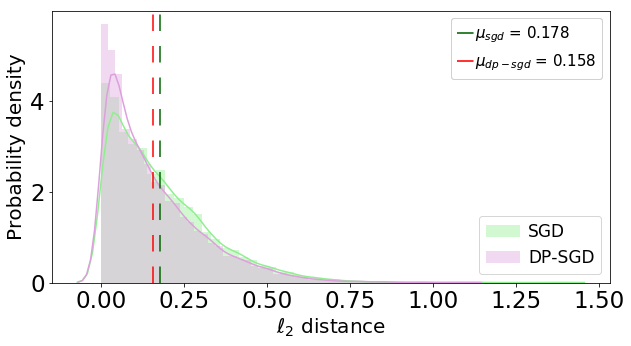}\label{deepcifar}}
  \hfill
  \subfloat[Curvature profiles as largest 10 eigenvalues of Hessian matrix for CIFAR-10. ]{\includegraphics[width=0.24\textwidth]{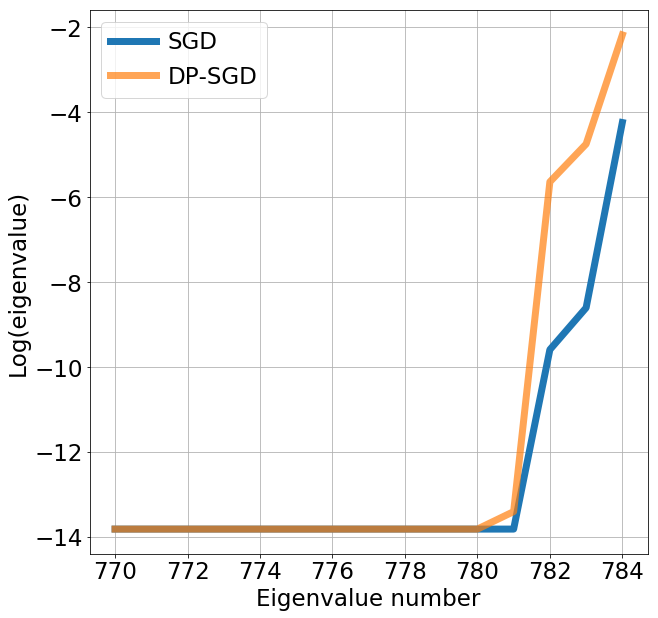}\label{curecifar}}
  \caption{Results of CIFAR-10 experiments.}

\end{figure*}

The standard non-private SGD in this settings reaches test accuracy $81.3\%$, while the DP-SGD with $\sigma=1$ and $C=3$ is able to reach the test accuracy $70.1\%$. The large gap $11.2\%$ between two models on the baseline not perturbed CIFAR-10 dataset makes comparison of the robustness with adversarially perturbed (FGSM, PGD) and commonly corrupted samples uninformative. In case of CIFAR-10, we measure and compare the robustness as the distance to the closest decision boundary and as a curvature profile.

In Figures \ref{deepcifarinf} and \ref{deepcifar}, the results for distances to the closest boundary are shown. As in Section \ref{deepfoolsection}, the boundaries for DP-SGD are closer to samples. With the addition of DP, the average value for $\ell_{\infty}-$bounded closest distance is reduced from $0.268$ to $0.238$ ($\sim 11.2\%$ drop) and $\ell_{2}-$bounded closest distance is reduced from $0.178$ to $0.158$ ($\sim 11.2\%$ drop). The smaller distance to the closest decision boundary might indicate that the DP-SGD model is more susceptible to DeepFool adversarial examples (less robust).

Similarly to Section \ref{curesection}, we compute the curvature as the largest $10$ eigenvalues of Hessian matrix with respect to input data at $100$ random test samples, and show the average curvature in Figure \ref{curecifar}. Although not all eigenvalues for DP-SGD are higher than SGD, the largest and the most important eigenvalue of the DP-SGD trained model is higher, which indicates that model trained with DP-SGD under these settings produces more curved decision boundaries, thus according to \cite{moosavi2019robustness} makes it less robust to adversarial examples. 

\section{Discussion and Related Work}
\label{relative}

From privacy point of view, some information is considered to be sensitive and secret, while an adversary targets to obtain that private information about the model's underlying training data or the model itself. Model privacy has been found to be violated, by showing the existence of several malicious approaches: restoring model details with model extraction attacks \cite{tramer2016stealing}, stealing the hyperparameters of the model \cite{wang2018stealing}, intentionally exploiting back-doors to the model to make it less private \cite{song2017machine}. Another type of privacy attacks, data privacy attacks, are using model and its outputs to recover sensitive information about the training data. One of them is the membership inference attack \cite{shokri2017membership}, where an attacker aims to identify whether a particular data point was a part of a training dataset or not. The threat from this type of attacks is clear in scenarios, such as health analytics, where the difference between case and control groups can leak patient’s personal conditions. Privacy attacks also include the learning global properties of private training data \cite{ganju2018property}, inferring missing information using a model and incomplete information \cite{fredrikson2014privacy}, obtaining meaningful data from models \cite{ateniese2013hacking}. Another important privacy attack is the model inversion attack \cite{fredrikson2015model}, where an adversary aims to approximate some of the model's training data points, using some intermediate layers output, or even using only outputs of the model. In \cite{fredrikson2015model}, the model inversion attack has been applied to a face classification model, that allowed to reconstruct recognizable faces from training data. 


To mitigate these privacy issues, several approaches to defend private information were proposed \cite{shokri2015privacy,hayes2018contamination,nasr2018machine,abadi2016deep}. In \cite{nasr2018machine}, the adversarial regularization was proposed to protect membership privacy. In \cite{shokri2015privacy,hayes2018contamination}, authors proposed to secure multi-party machine learning systems by the adversarial training on contamination attacks \cite{hayes2018contamination} or paralleling the optimization algorithms \cite{shokri2015privacy}. Abadi et al. \cite{abadi2016deep} proposed to apply the differential privacy \cite{dwork2008differential} to the learning procedure, which efficiently bounds the influence of each input data point to a mechanism's output. Differential privacy has been used in many classical machine learning algorithms to reduce privacy leakage of the training data \cite{chaudhuri2011differentially,kifer2012private,bassily2014private,song2013stochastic,dwork2010boosting,chaudhuri2013near,rubinstein2009learning,zhang2012functional}. 


Though this powerful technique successfully protects data samples from privacy attacks, as shown in \cite{abadi2016deep}, the accuracy of differentially private models is less in comparison to undefended models. Moreover, Bagdasaryan et al. \cite{bagdasaryan2019differential} recently discovered that it has significant impact on model's accuracy, by showing that reduction of accuracy and gaining in privacy are not borne equal. They showed that models trained with DP-SGD not only have less accuracy, but also have significantly less fairness for underrepresented classes. Authors demonstrated that DP-SGD trained gender classification model exhibits much lower accuracy for black people compared to white people, which identifies the model's additional unfairness. The intuitive explanation is the following: as underrepresented groups imply larger gradients, thus clipping and random noise reduce their influence on the model.

Several other works used concept of privacy and adversarial robustness in their studies \cite{lecuyer2019certified,du2019robust,jia2019memguard,song2019privacy,hayes2020provable,phan2019preserving}.
Lecuyer et. al \cite{lecuyer2019certified} proposed to use concepts of differential privacy to certify adversarial robustness of the model. They certify a prediction of the neural network with additive Gaussian noise to input data at test time and guarantee prediction label at some surrounding area for each sample. 
In \cite{du2019robust}, it was proposed to detect anomalies and backdoor poisoning attacks to a model via differential privacy. Authors train autoencoders in differentially private setting, by the similar per-example gradient clipping and noise addition, and use loss thresholds to detect anomalies and poisoning attacks.
In \cite{jia2019memguard}, a defensive methodology from the black-box membership inference attacks using adversarial examples was proposed. Their algorithm randomly adds noise to the confidence score vector predicted by the target classifier to transform a data sample into an adversarial example.
Similar to our work, but other way around, a research was recently done in \cite{song2019privacy}, where authors studied the unexpected privacy risks of securing models to adversarial examples. They showed that models trained with different adversarial training techniques are more vulnerable to the membership inference attack, which indicates that focusing solely on adversarial defense has impact on leakage of private information. Intuitively, they explain that adversarial defense strategy tries to robustify the model by making sure that outputs of $\ell_p$-ball around training data points remains unchanged, which leaks even more information about its membership during training process. 
Following \cite{song2019privacy}, theoretical analysis of the connection between the adversarial training and membership inference attacks was performed in \cite{hayes2020provable}. Authors proposed the theoretical and experimental framework to compare the standard and adversarially robust training methods from the membership privacy point of view. However, authors did not study 'reverse' problem of the effect of differential privacy to the robustness of the model.
Phan et al. \cite{phan2019preserving} proposed to use the adversarial training with differential privacy, but only briefly mentioned without proof the interplay trade-off between adversarial robustness and privacy of the model. We, on the other hand, experimentally show that differential privacy (under some settings) indeed reduces the model's robustness to adversarial examples and common corruptions.
Taking into consideration the related works above and observations of this paper (Section \ref{Experiments}), a principle question arises naturally:\newline
\newline
\centerline{\textit{\textbf{Is this trade-off a fundamental problem?}}}
\newline

It is hard to find a solid mathematical answer if differential privacy and robustness of a model are always necessarily in conflict. On one hand, there is no explicit connection between training models with differential privacy and decreasing model's robustness.

On the other hand, results of Section \ref{Experiments} empirically show that differential privacy (under some settings) decreases model's robustness to adversarial examples and common corruptions. Intuitive explanation for this effect is the following: training with differential privacy consists of clipping the individual per-example gradients and adding Gaussian noise to them for each of the update steps. It makes a model invariant to the existence of each sample in the training data, making impact of the samples to the model less observable. Therefore, differential privacy potentially tends to equalize (decrease or increase) the surrounding area (such as $\ell_p$ ball) of samples for which model prediction is unchanged, making overall the model less or more robust to different perturbations.

The difficulty of judgement the fundamentality of the privacy-robustness trade-off was also discussed in \cite{song2019privacy}, where authors empirically demonstrated the privacy risks of securing models against adversarial examples. The mathematical formulation of the trade-off between adversarial training and membership inference was done in \cite{hayes2020provable}. Authors proved that an adversarially trained model can either be more or less susceptible to the membership inference attack than a standard model, showing there exist settings with and without privacy-robustness trade-off. The settings are conditioned on $\ell_p-$ball radius $\varepsilon$ used in the adversarial training, and training size $n$.

\begin{wrapfigure}{r}{0.5\textwidth}
\vspace{-0.2cm}
  \centering
  \includegraphics[width=0.95\linewidth]{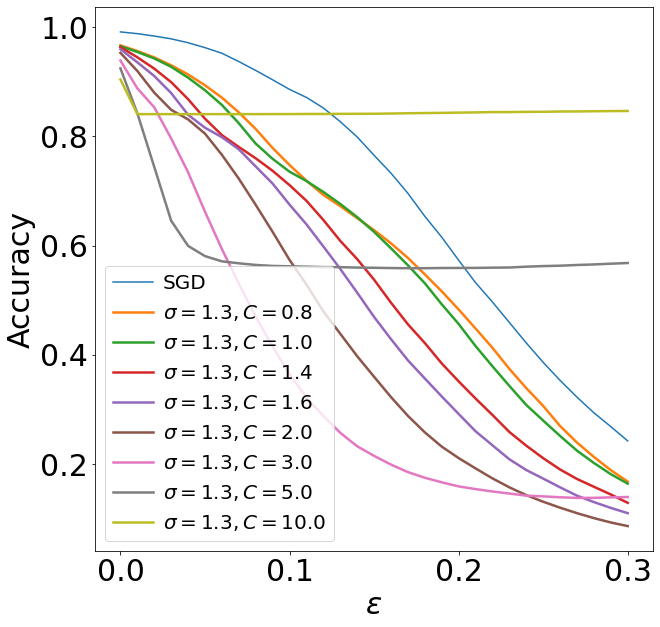}
  \caption{Dependence of FGSM accuracy curves on clipping bound for MNIST}
  \label{fig8}
  \vspace{-0.8cm}
\end{wrapfigure}

All the experiments in the Section \ref{Experiments} demonstrate settings, under which differential privacy decreases robustness of the model. However, in Figure \ref{noisb}, we see that for the sufficiently big clipping bound ($C=3.0$), there is an effect of 'plateau' in terms of adversarial robustness with strong FGSM adversaries (big magnitude of $\ell_\infty$). Increasing the magnitude of the clipping bound $C$ to even larger value, we unexpectedly got the surprising results: the 'plateau' accuracy line goes higher as the clipping bound $C$ increases. Figure \ref{fig8} demonstrates that surprising effect of settings when differential privacy helps improving robustness of the model to FGSM adversaries.

The results in Figure \ref{fig8} support findings in \cite{hayes2020provable}, that there exist some settings, when there is no conflict between privacy and robustness, and, moreover, improving one might also positively affect another. The increase of robustness with increasing $C$ is primarily due to the existence of $C$ in the gradient noise term (the gradient noise in DP-SGD is $\mathcal{N}(0, \sigma^2C^2\textbf{I})$). Gradient noise have been found to improve learning for very deep networks \cite{neelakantan2015adding}. Authors assert that gradient noise helps the model to explore the optimization landscape, escaping saddle points and local minima, possessing attractive robustness properties. 

\section{Conclusion}
It is well-known that neural networks trained with standard gradient descent methods, such as SGD, are brittle and not secure to different out-of-distribution examples. In this paper, we experimentally demonstrated that models, trained by differentially private SGD, that strongly protects the privacy of the model and underlying training data, might be even less secure, i.e. more susceptible to adversarial examples and common corruptions. In our experiments, we studied five different robustness measurements: accuracy drop after a single FGSM attack, robustness against PGD attack and a number of iterations to achieve adversarial class, distance to the closest decision boundary in $\ell_2$ and $\ell_{\infty}$, curvature profile of the decision boundary and performance on a corrupted dataset. We also studied how the main components of differentially private training (gradient clipping and noise addition) affect the robustness. The results indicate, that differential privacy reduces not only model's accuracy, but also affect the robustness of the model, decreasing or increasing. This threatens differential privacy to be deployed in security-critical scenarios. Interesting effect is observed with big gradient noise, when privacy and robustness are not in conflict. This effect and solid mathematical explanation of connection between differential privacy and robustness are the perspective directions for the future work.

\bibliographystyle{splncs04}
\bibliography{main}
\end{document}